\documentclass[journal]{IEEEtran}

\ifCLASSINFOpdf
\else
\fi

\usepackage[cmex10]{amsmath}
\usepackage[lined,boxed,commentsnumbered]{algorithm2e}

\usepackage{amssymb}
\usepackage{amsthm}
\newtheoremstyle{j}%
{3pt}%
{3pt}%
{}%
{\parindent}%
{\bfseries}%
{.}%
{.5em}%
{}%

\theoremstyle{plain}

\newtheorem*{rem*}{Remark}
\newtheorem*{concl*}{Conclusion}
\newtheorem*{theorem*}{Theorem}
\newtheorem*{cor*}{Corollary}
\newtheorem*{algo*}{Algorithm}

\newtheorem{theorem}{Theorem}[section]

\newtheorem{deff}[theorem]{Definition}

\theoremstyle{definition}

\newtheorem*{example*}{Example}

\newtheorem*{prob*}{Problem}

\newcommand{\norm}[1]{\left\lVert#1\right\rVert}

\usepackage{float}
\usepackage{graphicx}
\usepackage{tabularx}
\usepackage{color}
\usepackage{overpic}

  \usepackage[caption=false,font=footnotesize,labelfont=sf,textfont=sf]{subfig}

\newcommand{\Z}{\mathbb{Z}}

\newcommand{\R}{\mathbb{R}}

\newcommand{\psihat}{\widehat{\psi}}
\newcommand{\psitilde}{\widetilde{\psi}}

\newcommand{\eps}{\varepsilon}

\newcommand{\bal}{\begin{align}}
\newcommand{\eal}{\end{align}}

\newcommand{\bM}{\begin{pmatrix}}
\newcommand{\eM}{\end{pmatrix}}

\DeclareMathOperator{\diag}{diag}

\DeclareMathOperator{\argmin}{argmin}

\makeatletter
\newcommand{\onelettername}[1]{#1\aftergroup\@gobble}
\makeatother

\definecolor{B}{rgb}{0.15,0.4,0.8}
\definecolor{olive}{rgb}{0.3, 0.4, .1}
\definecolor{dgreen}{rgb}{0.,0.6,0.}
\definecolor{dred}{rgb}{0.9,0.1,0.1}

\begin{document}
\title{Shearlet-based compressed sensing for fast 3D cardiac MR imaging using iterative reweighting}

\author{Jackie Ma \thanks{J. Ma is with the   Image   and   Video   Coding
Group,   Fraunhofer   Institute   for   Telecommunications--Heinrich   Hertz   Institute,  Berlin  10587,  Germany  (e-mail:  jackie.ma@hhi.fraunhofer.de).}
\quad 
Maximilian M\"{a}rz \thanks{M. M\"{a}rz, and G. Kutyniok are with the Department
of Mathematics,   Technische Universit\"{a}t Berlin, Berlin, 10623 Berlin, Germany 
(e-mail: \{maerz, kutyniok\}@math.tu-berlin.de).}
\quad 
Stephanie Funk
\quad
Jeanette Schulz-Menger\thanks{S. Funk and J. Schulz-Menger are with the Working Group Cardiovascular Magnetic Resonance, Experimental and Clinical Research Center, a joint cooperation between the Charit\'{e} Medical Faculty and the Max-Delbrueck Center for Molecular Medicine; and HELIOS Klinikum Berlin Buch, Department of Cardiology and Nephrology, Lindenbergerweg 80, 13125 Berlin, Germany (e-mail: \{stephanie.funk, jeanette.schulz-menger\}@charite.de).}
\quad
Gitta Kutyniok 
\quad

Tobias Schaeffter
\quad
Christoph Kolbitsch \thanks{T. Schaeffter and C. Kolbitsch are with the Physikalisch-Technische Bundesanstalt (PTB), Braunschweig and Berlin, Germany and Division of Imaging Sciences and Biomedical Engineering, King'{}s College London, London, UK (e-mail: \{tobias.schaeffter, christoph.kolbitsch\}@ptb.de).

This work has been submitted to the IEEE for possible publication. Copyright may be transferred without notice, after which this version may no longer be accessible.}
}

\markboth{IEEE Transactions on Medical Imaging}%
{Shell \MakeLowercase{\textit{et al.}}: Bare Demo of IEEEtran.cls for Journals}

\maketitle

\begin{abstract}
\noindent
High-resolution three-dimensional (3D) cardiovascular magnetic resonance (CMR) is a valuable medical imaging technique, but its widespread application in clinical practice is hampered by long acquisition times. Here we present a novel compressed sensing (CS) reconstruction approach using shearlets as a sparsifying transform allowing for fast 3D CMR (3DShearCS). Shearlets are mathematically optimal for a simplified model of natural images and have been proven to be more efficient than classical systems such as wavelets. Data is acquired with a 3D Radial Phase Encoding (RPE) trajectory and an iterative reweighting scheme is used during image reconstruction to ensure fast convergence and high image quality. In our in-vivo cardiac MRI experiments we show that the proposed method 3DShearCS has lower relative errors and higher structural similarity compared to the other reconstruction techniques especially for high undersampling factors, i.e. short scan times. In this paper, we further show that 3DShearCS provides improved 
depiction of 
cardiac anatomy (measured by assessing the sharpness of coronary arteries) and two clinical experts qualitatively analyzed the image quality.  
\end{abstract}


\IEEEpeerreviewmaketitle

\section{Introduction}
\IEEEPARstart{M}{}agnetic Resonance Imaging (MRI) is a valuable medical imaging technique that can capture complex anatomical structures functionalities and is also established in differentiating tissues. In contrast to this, coronaries are still very challenging. Further, MRI allows for three-dimensional (3D) imaging with high spatial resolution. This is especially important for cardiac applications such as the assessment of coronary arteries or in congenital heart disease to visualize complex anatomical structures. 

The main challenge of high-resolution 3D cardiac MRI are long acquisition times. In order to achieve the necessary spatial resolution, MR data is obtained during free-breathing over multiple respiratory and cardiac cycles. Data acquisition is restricted to predefined respiratory (e.g. end-expiration) and cardiac (e.g. mid-diastole) phases to minimize motion artefacts. This approach provides excellent image quality but can lead to scan times of more than 15 min which makes it challenging to apply in clinical practice \cite{kolbitsch_magn_reson_med_03_2011, pang_mrm_2014}.

Several approaches have been proposed to reduce scan times by acquiring less data and utilizing additional information about the acquired data in the image reconstruction. Parallel imaging techniques for example use the spatial information from multiple receiver coils to improve the conditioning of the image reconstruction problem \cite{sodickson_magn_reson_med_10_1997, pruessmann_magn_reson_med_11_1999, Griswold2002}. Further improvements in image quality can be achieved by applying the framework of compressed sensing to MR image reconstruction \cite{Lustig2008}. If the MR image itself or a transformation of the MR image is sparse, this sparsity can be used to suppress undersampling artefacts and improve image quality.  

Commonly, wavelet transforms are used as sparsity transforms for 3D anatomical MR images. Although this approach is very robust for a wide range of different images and applications, the wavelet transform is not necessarily optimal to accurately describe anatomical images of the heart. Recently, the shearlet transform has been proposed for medical image reconstruction \cite{Ma,KutLimCS}. The shearlet transform is based on a multiscale directional system that provides mathematically provable optimal approximation rates of so-called \emph{cartoon-like functions} \cite{KutLim} which are a simplified model for images, in particular, medical images. Wavelets do not fulfill such an optimal approximation rate of curvi-linear singularities which represent the edges in an image. This favors the choice of shearlets in image reconstruction problems such as denoising and inpainting. The two main reasons why shearlets outperform wavelets in terms of the approximation rate of images is that shearlets are build upon 
anisotropic scaling and shearing. The latter allows the elements to have different directionalities. These two properties allow for elongated and directional 
elements 
which are much more adapted to curves than isotropic and non directional elements such as for wavelets.

Previous studies have shown that shearlet-based compressed sensing (CS) approaches yield an improved MR image quality compared to wavelet-based reconstructions \cite{Ma, KutLimCS}. Nevertheless, so far these studies have only been carried out using simulated 2D MR data of the brain. Data acquisition in these simulations was assumed to be Cartesian (i.e. each k-space point is located on a uniform grid) which strongly simplifies the image reconstruction process. In addition, the spatially varying sensitivity of receiver coils commonly used to record the MR signal was not taken into consideration in these studies.

In this paper we present a non-Cartesian 3D shearlet-based CS reconstruction approach which overcomes these limitations and allows for the reconstruction of 3D high-resolution image from MR data obtained on arbitrary k-space locations using multiple receiver coils. In this case the reconstruction operator is not easily diagonalizable anymore and requires an iterative solver. In order to ensure high image quality and achieve fast convergence, which is especially important for medical in-vivo applications, an adapative reweighting procedure is used \cite{MaMae}. The proposed 3D shearlet-based CS approach is evaluated on in-vivo cardiac MR scans of five healthy volunteers. Image quality is assessed using general image quality metrics and clinical diagnostic scores.

\section{Methods}

\subsection{MR acquisition model}
In MRI, data is acquired in Fourier-space or k-space and the transform of the obtained k-space data $y$ to the image data $x$ can be described with the encoding operator $E$ 
\begin{align}
y =  Ex = GFSx .
\end{align}
$S$ describes the spatial distribution of the sensitivity of the receiver coils which are used to record the MR signal. It can be calculated from an additional calibration scan or from the image data itself. $F$ is the Fourier operator and is commonly realized using a discrete fast Fourier transform (FFT). If the obtained k-space points are located on a Cartesian grid, $G$ is simply a mask selecting the acquired data. For non-Cartesian acquisition schemes, $G$ requires the interpolation of data points from arbitrary positions onto a cartesian grid \cite{jackson_tmi_1991}.

\subsection{Compressed sensing}

\emph{Compressed Sensing (CS)} has been introduced by Donoho in \cite{Don} and Cand\`{e}s, Romberg, and Tao in \cite{CanRomTao1, CanRomTao2} as a methodology that allows for successful reconstructions using considerably less information than other conventional methods such as the reconstruction formula given by the Shannon-Nyquist theorem \cite{Sha}. One of the key ideas is to incorporate randomness into the measurement process and solving an $\ell^1$-minimization problem of the form
\begin{align}
\min_x \| x \|_1 \quad \text{subject to} \quad \| y- Ex \|_2 \leq \eps, \label{eq:l1synth}
\end{align}
where $E$ denotes the \emph{encoding matrix} that gives rise to the \emph{measured data} $y$ with an \emph{estimated accuracy} $\eps>0$.
Problem \eqref{eq:l1synth} is known to have a unique sparse solution provided $E \in \R^{m\times N}$ satisfies certain properties such as the so-called \emph{Restricted Isometry Property} \cite{Can}. Moreover, one of the key assumptions is that the vector of interest $x \in \R^N$ is
\emph{sparse}, i.e. the number of non-zero elements
\begin{align*}
 \#\{ k \in \{ 1, \ldots, N\} \, : \, x_k \neq 0 \}
\end{align*}
is small compared to $N$. However, in many applications the object of interest $x$ is not directly sparse but only after the application of a suitable transform $\Psi$. This is, for example, the case in MR imaging. Hence, instead of solving \eqref{eq:l1synth}, Lustig et al. have considered
\begin{align}
\min_x \| \Psi x \|_1 \quad \text{subject to} \quad \| y- Ex \|_2 \leq \eps \label{eq:l1ana}
\end{align}
in \cite{Lustig2008} for recovering MR data from highly undersampled data and a sparsifying transform $\Psi$. There are many different possible transforms for $\Psi$, with the \emph{wavelet transform} \cite{Dau} being the most prominent one. Despite its common use, wavelet transforms do not necessarily provide an optimal description of MR images. As we have already mentioned in the introduction a multiscale directional transform, such as the  \emph{shearlet transform} \cite{Lim1, Lim2} can mathematically be shown to outperform the classical wavelet transform \cite{ShearletBook} in terms of its approximation rate. The approximation using wavelets and shearlets for MRI using uniform samples is discussed in \cite{Ma}. 

In this work  we will use an algorithm developed in \cite{MaMae} together with the shearlet transform. The algorithm is not solely build for the shearlet transform but more generally for multilevel transform, which will be discussed in Section \ref{sec:MW} and Section \ref{sec:algo}. In Section \ref{sec:Shearlets} we will then present the 2D and 3D shearlet system known from the literature as both systems will be examined in this work.

\subsection{Multilevel reweighting} \label{sec:MW}

The idea of using reweighting in order to improve solutions of \eqref{eq:l1synth} was first considered by Cand\`{e}s et al. in \cite{CanWakBoy}. In order to enhance the sparsity of the recovered coefficients a \emph{weighting matrix} $W = \diag(\sigma_1, \ldots, \sigma_N) = (\sigma_i \delta_{i,j})_{i,j \leq N}$ is introduced, where $\delta_{i,j} =1$ if $i =j$ and $0$ otherwise. Further, the weights $\sigma_i>0, i = 1, \ldots, N$ are adapted to the sparsity structure of the object of interest $x^*$. More precisely, suppose $x^* = (x^*_1, \ldots, x^*_N) \in \R^N$ is the true signal that one wishes to recover. Then the minimization problem
\begin{align}
\min_x \| Wx \|_1 \quad \text{subject to} \quad  \|y - E\|_2 \leq \eps \label{eq:l1synthW}
\end{align}
with weights
\begin{align}
\sigma_i = \begin{cases} \frac{1}{|x^*_i|}, & x_i, \neq 0 \\ \infty,& \text{otherwise},\end{cases} \label{eq:weightsSyn1}
\end{align}
will ideally find sparser solutions than the unweighted minimization problem in \eqref{eq:l1synth} \cite{CanWakBoy}. Note that the weights shown in \eqref{eq:weightsSyn1} are practically not feasible as they already assume the knowledge of the true signal $x^*$ that one wishes to recover. Hence, one usually considers a sequence of weights
\begin{align}
\sigma_i^k = \frac{1}{|x_i^k| + \nu}, \label{eq:weightsSyn}
\end{align}
where $\nu>0$ is small and $(x^k)_k \subset \R^N$ is a sequence of approximations to the true signal $x^*$ obtained by 
\begin{align*}
x^k = \argmin_x \| W^{k-1}x \|_1 \quad \text{subject to} \quad   \|y - Ex\|_2 \leq \eps.
\end{align*}
Note that the weights defined in \eqref{eq:weightsSyn} penalize small coefficients stronger and thus contribute the information that they are more likely zero in the true signal. 
When adopting this idea to the minimization problem considered in \eqref{eq:l1ana}, the straightforward implementation would yield weights of the form
\begin{align}
\sigma_i^k =  \frac{1}{|(\Psi x^k)_i| + \nu}, \label{eq:weightsAna}
\end{align}
where
\begin{align*}
x^k = \argmin_x \| W^{k-1} \Psi x \|_1 \quad \text{subject to} \quad   \|y - Ex\|_2 \leq \eps.
\end{align*}
Although the coefficients computed from a multiscale transform can be sparse, they intrinsically decrease due to the multiscaling. In order to take full advantage of the reweighting scheme explained above, the natural decrease of the coefficients in magnitude has to be compensated for in order to avoid wrong classifications of zero coefficients. 

Recently, in \cite{MaMae} we have proposed the use of weights that are associated with levels  $j \in \{1, \ldots, J\}$ that correspond to the scales of the multiscale transform such as the wavelet transform and shearlet transform.  The general idea of multilevel reweighting is now not to equally apply the classical reweighting across all levels but rather separately within each level. The resulting proposed weights are then
\begin{align}
\sigma_{i_j}^k = \frac{\lambda_j}{ |(\Psi x^k)_{i_j}| + \nu}, \quad i_j \in \{ N_{j-1}+1, \ldots, N_{j} \} \label{eq:multiweightsAna}
\end{align}
per level $j$ and index sets $\{ N_{j-1}+1, \ldots, N_{j} \}$ that partition the levels, i.e., 
$$
\{1, \ldots, N\} = \bigcup_{j =1}^J\{ N_{j-1}+1, \ldots, N_{j} \}
$$
This reweighting technique has been shown to achieve faster convergence and yield improved final image quality for 2D images \cite{MaMae}. Note that this method is computationally very demanding especially for 3D images as the weights have to be stored and computed after each iteration.

\subsection{Algorithm}\label{sec:algo}

The iterative reweighting can be directly incorporated into the \emph{alternating direction method of multipliers (ADMM)} which is a key tool to solve sparse regularized inverse problems such as the MR image reconstruction problem that we consider in this paper. For ADMM, consider
\begin{align*}
\min_x \norm{\Psi x}_1 + \frac{\beta}{2} \norm{y- Ex}_2^2,
\end{align*} 
which is equivalent to \eqref{eq:l1ana} for a suitably chosen parameter $\beta$. Then the resulting ADMM  steps for solving the latter problem are
\begin{align*}
x^{k+1} & = \argmin_x \frac{\beta}{2} \norm{y- Ex}_2^2  + \frac{\mu}{2} \|d-\Psi x -b^k\|_2^2, \\
d^{k+1} & = \argmin_d \norm{d}_1 + \frac{\mu}{2} \|d-\Psi x^{k+1} - b^k\|_2^2, \\
b^{k+1} & = b^k + \Psi x^{k+1} - d^{k+1}, 
\end{align*}
for a parameter $\mu>0$. In order to solve the $x$-update, the system 
\begin{align}
\label{eq:system}
\left(\beta E^* E + \mu \Psi^ * \Psi \right) x = \beta E^*y + \mu \Psi^ *(z^k - u^k)
\end{align}
has to be solved. Equation \eqref{eq:system} is now a key equation and its solvability depends strongly on the matehmatical properties of the operators. In fact, if the encoding operator $E$ is simply given as a subsampled discrete Fourier transform, the matrix on the left hand side is diagonalizable by the discrete Fourier transform and therefore the system \eqref{eq:system} is explicitly solvable in $\mathcal{O}(n \log n)$ flops. 

For our non-Cartesian encoding operator $E$ however, we propose to use an iterative method to solve the system approximately. It is known that ADMM still converges, although \eqref{eq:system} is not solved up to full precision \cite{Deng2016}. By using the solution of the previous iterate one can use a \emph{warm start} so that only a few iteration steps of the iterative method are necessary for the entire algorithm to converge. %

Note that by solving the $d$-problem we obtain the sparsifying transform coefficients. The idea of using the weights to obtain even sparser transform coefficients can be directly incorporated in this subproblem. Hence, following \cite{MaMae}, the multilevel adapted iterative reweighting steps are directly incorporated into the $d$-update as follows:
\begin{align}
d^{k+1} = \argmin_d \|W^{k+1} d\|_1 + \frac{\mu}{2} \|d-\Psi x^{k+1} - b^k\|_2^2,
\label{eq:prox}
\end{align} 
where the weighting matrix is given as in \eqref{eq:multiweightsAna} by
\begin{align}
\sigma_{i_j}^{k+1} = \frac{\lambda_j}{ |(\Psi x^{k})_{i_j}| + \nu}, \quad i_j \in \{ N_{j-1}+1, \ldots, N_{j} \},
\end{align}
and $W^{k+1} = \diag (\sigma_1^{k+1},\dots,\sigma_n^ {k+1})$. The proximal step \eqref{eq:prox} is then explicitly solved by
\begin{align*}
d^{k+1} = \text{shrink} \left(\Psi x^{k+1} + b^k , \frac{1}{\mu}W\right),
\end{align*}
where 
\begin{align*}
\text{shrink}(z,\lambda) = \begin{cases} \max (\norm{z} - \lambda,0) \frac{z}{\norm{z}}, & z\neq 0\\ 0, & z = 0 \end{cases}
\end{align*}
is applied element-wise. 

\subsection{Shearlets}\label{sec:Shearlets}

Shearlet systems were first introduced by Labate, Lim, Guo Kutyniok and Weiss in \cite{LLKW2007, Lim1, GuoKutLab2006} as a directional representation system that provably outperforms classical systems such as wavelets within a certain model. The model assumption is that the function that is to be approximated is a so-called \emph{cartoon-like function}. These are functions that are smooth up to a smooth discontinuity curve. These functions serve as a model for natural images. It is known that wavelets, for instance, can only provide a best $N$-term approximation rate of the order $N^{-1}$ while shearlets do, up to log factors, reach $N^{-2}$ \cite{ShearletBook,KutLim} which is optimal for this class of functions \cite{DonSparse}. Moreover, similar to wavelets a construction using compactly supported generators can also be obtained which in turn allows fast implementations \cite{KitKL2012}.

The novelty of this system compared to classical wavelet systems are the parabolic scaling matrices 
\begin{align*}
	A_{2^j} = \begin{pmatrix} 2^j & 0 \\ 0 & 2^{j/2 } \end{pmatrix}, 
	\qquad \widetilde{A}_{2^j} = \begin{pmatrix} 2^{ j/2 } & 0 \\ 0 & 2^j \end{pmatrix}, \quad j \geq 0 
\end{align*}
and the shear matrix
\begin{align*}
	S_k = \begin{pmatrix} 1 & k \\ 0 & 1 \end{pmatrix}, \quad  k \in \Z.
\end{align*}
These matrices are used in the following definition of a \emph{shearlet system}.
\begin{deff}[\cite{KitKL2012}]\label{Definition:ShearletSystem}
Let $\phi, \psi, \psitilde \in L^2(\R^2)$ be the \emph{generating functions} and $c=(c_1, c_2) \in \R^+ \times \R^+$. Then the
\emph{(cone adapted discrete) shearlet system} is defined as
\begin{align*}
	\mathcal{SH}(\phi,\psi, \psitilde, c) = \Phi(\phi, c_1) \cup \Psi (\psi, c) \cup \widetilde{\Psi}(\psitilde,c),
\end{align*}
where
\begin{align*}
	\Phi(\phi,c_1) &= \{ \phi( \cdot - c_1m) \, : \, m \in \Z^2\},\\
	\Psi (\psi, c) &= \left\{ \psi_{j,k,m}  \, 	: \, j \geq 0, |k| \leq 2^{j/2}, m  \in \Z^2 \right\},\\
	\widetilde{\Psi} (\psitilde, c) &= \left\{ \psitilde_{j,k,m}\, : \, 	j \geq 0, |k| \leq 2^{j/2}, m  \in \Z^2 \right\},
\end{align*}
and
\begin{align*}
	\psi_{j,k,m} &= 2^{3j/4 } \psi \left(( S_k A_{2^j}) \cdot - cm \right),  \\
	\psitilde_{j,k,m} &= 2^{3j/4 } \psitilde \left(( S_k^T \widetilde{A}_{2^j}) \cdot - \widetilde{c}m \right) .
\end{align*}
The multiplication of $c$ and $\widetilde{c} = (c_2,c_1)$ with the translation parameter $m$ should be understood 
entry wise. 
\end{deff}

The attentive reader might wonder, why a shear action is used to obtain a directional component and not, for instance, rotation. Indeed, rotation has been used before shearing yielding the well-known \emph{curvelets} by Cand\`{e}s et al. \cite{CanDon}. However, rotation does not leave the integer grid invariant which is particularly desired when implementing these systems.

Definition \ref{Definition:ShearletSystem} concerns the case of 2D shearlets. Although we will also use these systems in this paper, our main purpose of study is 3D data. For such, one could use 2D shearlets along slices. However, as we will show, 3D shearlets yield better results from numerous different perspectives. The definition of 3D shearlets is a straightforward generalization of 2D shearlets \cite{3DShearlets, KutLimRei} and we shall only give a brief presentation in this paper. The scaling matrices used for the 3D system are
\begin{align*}
	A_{2^j} &= \begin{pmatrix} 2^j & 0 & 0\\ 0 & 2^{j/2 } & 0 \\ 0 & 0 & 2^{j/2} \end{pmatrix}, 
	\quad \widetilde{A}_{2^j} = \begin{pmatrix} 2^{j/2} & 0 & 0\\ 0 & 2^{j } & 0 \\ 0 & 0 & 2^{j/2} \end{pmatrix},\\
	 &\qquad \qquad  \widehat{A}_{2^j} = \begin{pmatrix} 2^{j/2} & 0 & 0\\ 0 & 2^{j/2 } & 0 \\ 0 & 0 & 2^{j} \end{pmatrix},
	\quad j \geq 0, 
\end{align*}
and the shear matrices are
\begin{align*}
	S_{k} &= \begin{pmatrix} 1 & k_1 & k_2\\ 0 & 1 & 0 \\ 0 & 0 & 1 \end{pmatrix}, 
	\quad \widetilde{S}_{k} = \begin{pmatrix} 1 & 0 & 0\\ k_1 & 1 & k_2 \\ 0 & 0 & 1 \end{pmatrix},\\
	 &\qquad \qquad  \widehat{S}_{k} = \begin{pmatrix} 1 & 0 & 0\\ 0 & 1 & 0 \\ k_1 & k_2 & 1 \end{pmatrix},
	\quad k_1, k_2 \in \Z. 
\end{align*}
Using these $3 \times 3$ matrices once can extend the idea of cone adapted shearlets to the so-called \emph{pyramid adapted shearlets}.
\begin{deff}[\cite{3DShearlets}]\label{Definition:ShearletSystem3D}
Let $\phi, \psi, \psitilde, \psihat \in L^2(\R^3)$ be the \emph{generating functions} and $c=(c_1, c_2, c_2) \in \R^+\times \R^+ \times \R^+$. Then the
\emph{(pyramid adapted discrete) shearlet system} is defined as
\begin{align*}
	\mathcal{SH}(\phi,\psi, \psitilde, c) = \Phi(\phi, c_1) \cup \Psi (\psi, c) \cup \widetilde{\Psi}(\psitilde,c) \cup \widehat{\Psi}(\psihat,c),
\end{align*}
where
\begin{align*}
	\Phi(\phi,c_1) &= \{ \phi( \cdot - c_1m) \, : \, m \in \Z^3\},\\
	\Psi (\psi, c) &= \left\{ \psi_{j,k,m}  \, 	: \, j \geq 0, |k_1|,|k_2| \leq 2^{j/2}, m  \in \Z^3 \right\},\\
	\widetilde{\Psi} (\psitilde, c) &= \left\{ \psitilde_{j,k,m}\, : \, 	j \geq 0, |k_1|,|k_2| \leq 2^{j/2}, m  \in \Z^3 \right\},\\
	\widehat{\Psi} (\psihat, c) &= \left\{ \psihat_{j,k,m}\, : \, 	j \geq 0, |k_1|,|k_2| \leq 2^{j/2}, m  \in \Z^3 \right\},
\end{align*}
and
\begin{align*}
	\psi_{j,k,m} &= 2^{j } \psi \left(( S_k A_{2^j}) \cdot - cm \right), \\
	\psitilde_{j,k,m} &= 2^{j } \psitilde \left(( \widetilde{S}_k \widetilde{A}_{2^j}) \cdot - \widetilde{c}m \right), \\
	\psihat_{j,k,m} &= 2^{j } \psitilde \left(( \widehat{S}_k \widehat{A}_{2^j}) \cdot - \widetilde{c}m \right) .
\end{align*}
The multiplication of $c$ and $\widetilde{c} = (c_2,c_1)$ with the translation parameter $m$ should be understood 
entry wise. 
\end{deff}

\section{Experiments}
The shearlet-based CS image reconstruction approach was  assessed in 3D in-vivo cardiac MR images. The performance of CS-based image reconstruction schemes depends strongly on the undersampling properties of the obtained MR data. To ensure high image quality, MR data has to be acquired in a way such that undersampling artefacts lead to incoherent signal contributions. Therefore, experiments were carried out with a radial phase encoding (RPE) MR sampling scheme. RPE has been shown previously to provide 3D high-resolution images even for high undersampling factors \cite{Boubertakh2009, kolbitsch_magn_reson_med_03_2011}. The RPE sampling scheme also allows for retrospective undersampling and simulation of different MR scan times. 

\subsection{MR Data Acquisition}
3D whole-heart MR data was acquired with a balanced steady state free precession sequence to ensure optimal MR signal strength. Fat suppression and a T2 preparation pulse (TE = 50 ms) was applied to enhance image contrast. Sequence parameters were: field of view of 288 x 288 x 288 $mm^3$ with an isotropic resolution of 1.5 $mm^3$, flip angle of $90^o$ and repetition/echo times of 4.3/2.2 ms. 

Along each RPE line a partial Fourier factor of 0.75 was applied and 64 RPE lines were obtained leading to an undersampling factor of 4 compared to a fully sampled Cartesian acquisition. Image reconstruction was carried out offline using Matlab (The MathWorks, Inc., Natich, MA, USA). The coil sensitivity information was calculated from the data itself and homodyne weighting was used to compensated for the partial Fourier acquisition \cite{kolbitsch_magn_reson_med_03_2011}. 

The obtained data was retrospectively undersampled by a factor of 1, 2, 4 and 6 leading to images with a total undersampling factor (R) of 4, 8, 16 and 24 equivalent to average scan times of 12.6 min, 6.3 min, 3.2 min and 1.6 min. 

Each data set was reconstructed using 
\begin{itemize}
	\item a non-Cartesian iterative SENSE reconstruction technique (itSENSE) \cite{pruessmann_magn_reson_med_10_2001}
	\item a non-Cartesian iterative SENSE approach with a spatial total-variation constraint (TV) \cite{cruz_mrm_2015}
	\item a 3D wavelet-based CS method (WaveCS) \cite{MaMae}
	\item the proposed 3D iteratively reweighted shearlet-based CS approach (3DShearCS) \cite{MaMae}
\end{itemize}
WaveCS was implemented in the same framework as \text{3DShearCS} to correspond to classical wavelet based MR image reconstructions using a decimated 3D wavelet transform without adaptive reweighting but classical $\ell^1$-minimization as proposed in \cite{Lustig2008}.

\subsection{In Vivo Experiments}
Five healthy volunteers were imaged on a 1.5T MRI scanner (Philips Medical Systems, Best, The Netherlands) using a 32-channel cardiac phased array coil. Respiratory phase ordering with automatic window selection with a bin with of 3 mm and cardiac end-diastolic triggering were used to minimize respiratory and cardiac motion artefacts, respectively \cite{kolbitsch_magn_reson_med_03_2011}. Written informed consent was obtained from all participants in accordance with the ethical rules of our institution. 

\subsection{Evaluation of 3D Shearlet System}
In order to assess the performance of a 3D shearlet system compared to a 2D shearlet system we reconstructed images for R = 4, 8, 16 and 24 using a 2D shearlet system as described in Eq. \ref{Definition:ShearletSystem}  (2DShearCS). The 2D shearlet system was applied slice by slice in the transverse plane during a CS image reconstruction for one volunteer.
 
In order to successfully apply the proposed approach in clinical practice, reconstruction times also need to be kept at a minimum. The reconstruction times of the shearlet CS method are strongly determined by the total number of iterations and how often the weights are recalculated. We therefore evaluated the convergence of the reconstruction algorithm and the relative change of the weights for different iteration numbers.

\subsection{Evaluation of Image Quality}
We used two image error measurements namely the \emph{relative error} and the \emph{Haar wavelet-based perceptual similarity index (HaarPSI)}, \cite{HaarPSI}. 

The relative error is a standard measure and can be calculated using
\begin{align*}
 \frac{ \|x_{\text{ref}} -  x_{\text{rec}}\|_2}{\|x_{\text{ref}}\|_2},
\end{align*}
where $x_{\text{ref}}$ is the vectorized reference image, $x_{\text{rec}}$ is the vectorized reconstruction. Its value is dependent on the image content, making it difficult to compare between different volunteers with varying anatomy. Obviously a smaller number is preferred for the relative error and the relative error can easily be computed for 3D images. 

The HaarPSI on the other hand is supposed to simulate the human similarity perception of objects in 2D and is therefore calculated on individual 2D slices. Hence, we have averaged the HaarPSI along all slices along the left-right direction. This choice is arbitrary and could be replaced by any other slice-direction. For HaarPSI the algorithm computes Haar wavelet coefficients to determine local similarities between two input images. The resulting index is a number between zero and one where a larger numbers represent a stronger similarity between two images. In \cite{HaarPSI} the method has been tested intensively on data bases where images have been scored by humans and in almost all cases HaarPSI has outperformed other common state of the art similarity indices. We refer the interested reader to that work and the references therein.

The quality of the obtained images was also assessed by measuring the sharpness (VS) of the coronary arteries. The coronary arteries are small and complex structures with a diameter of approximately 3 mm \cite{piccini_radiology_2014} and are highly susceptible to undersampling artefacts. Therefore, they provide a sensitive metric on the quality of the reconstructed images. VS is calculated as the mean intensity of the vessel edges relative to the maximum intensity in the center of the vessel. A value of 1 corresponds to a very well defined vessel and a value of 0 means that the vessel could not be distinguished from the surrounding tissue, i.e. is not visible anymore. The assessment was carried out semi-automatically with a commercial tool developed for MR angiography \cite{etienne_magn_reson_med_10_2002}.

In addition, the in-vivo MR images were assessed by two clinical experts. They assessed if the images were of diagnostic quality (i.e. if the coronary arteries were visible), scored the images on a 3-point score (0 non-diagnostic; 1 good; and 2 excellent) and also selected the image which fitted best to a reference image for each undersampling factor and volunteer separately. The whole assessment was performed with blinded reconstructions, meaning the two experts did not know which reconstruction corresponded to which method. If several images were equally comparable to the reference, multiple nominations were used. The applied scoring method is also applied in \cite{Prothmannetal}.

All evaluations were carried out relative to a reference image. For in-vivo acquisitions it was not possible to obtain a fully sampled k-space due to long acquisition time not feasible in practice. Therefore, images reconstructed with itSENSE from the original data with R = 4 were used as reference images. The evaluations were carried out using Matlab and R and statistical significance was determined with a two sampled t-test with a p-value smaller than 0.05 considered statistically significant. 

\section{Results}

\subsection{Performance of 3D Shearlet Image Reconstruction}

In this section we provide a comparison between the proposed 3DShearCS and 2DShearCS. In the latter case the reconstructions are obtained slice by slice along the foot-head-direction of the 3D data. 2DShearCS does not utilize any information along the foot-head direction and thus does not take advantage of the full 3D data. We also analyze the convergence properties and the behavior of the weights for the proposed method 3DShearCS.

\begin{figure*}[h!]
\centering
\qquad \qquad  \quad
\begin{overpic}[width=0.9\textwidth]{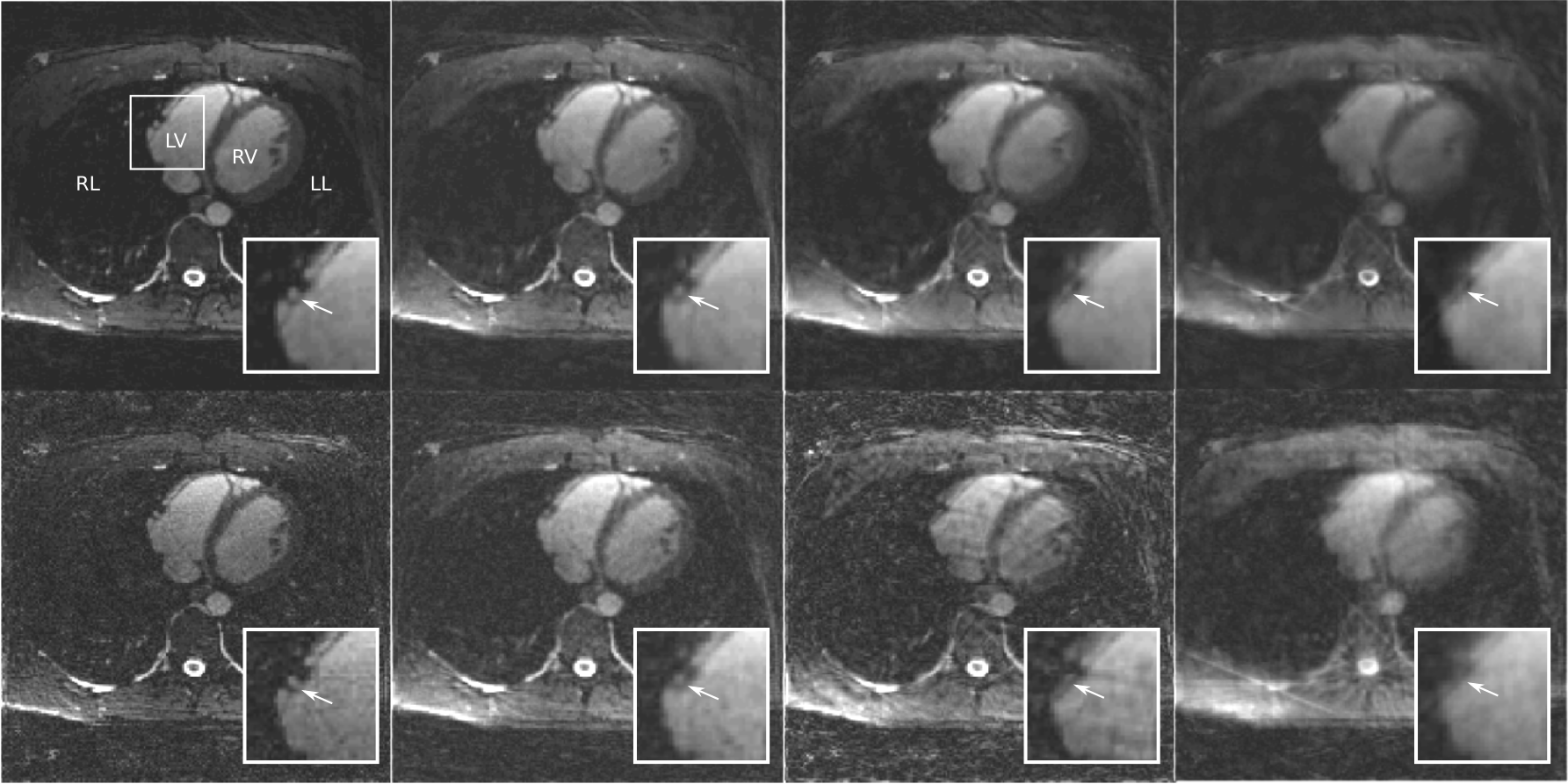} 
 \put(10,51){R = 4}
 \put(35,51){R = 8}
 \put(60,51){R = 16}
 \put(84,51){R = 24}
 \put(-11,36.5){3DShearCS}
 \put(-11,11.5){2DShearCS}
\end{overpic}
\caption{A reconstructed slice along the foot-head-direction using 3DShearCS and 2DShearCS for different undersampling factors R = 4, 8, 16, and 24. Small anatomical features such as the cross-section through the right coronary artery (white arrow) are still visible for R = 24 using 3DShearCS but are difficult to distinguish from undersampling artefacts with 2DShearCS even for R = 16. RV right ventricle, LV left ventricle, RL right lung, LV left lung.}\label{fig:NumSim}
\end{figure*}

Figure \ref{fig:NumSim} shows the comparison between 3DShearCS and the slice by slice 2DShearCS reconstruction. The 3D approach removes undersampling artefacts better than the 2D method. This becomes especially evident for higher undersampling factors. The additional degree of freedom of the 3D shearlet system allows for a better distinction between undersampling artefacts and anatomical structures and hence lead to a higher image quality. 

\begin{figure*}
\centering
 \includegraphics[width=\textwidth]{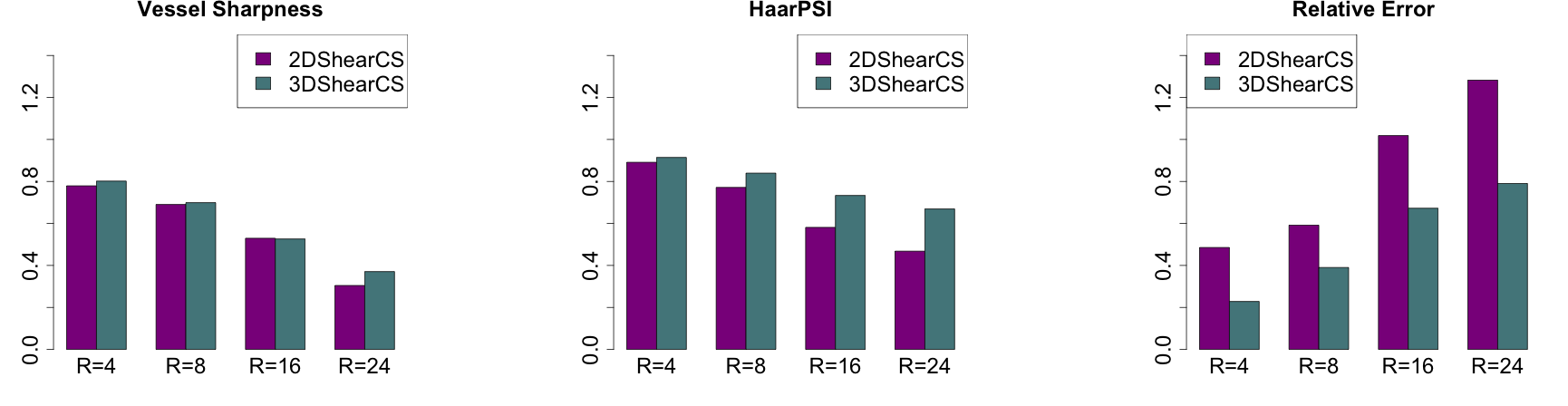}
\caption{Vessel sharpness, HaarPSI, and relative error for image reconstructions using 3DShearCS and 2DShearCS for undersampling factors R = 4, 8, 16, and 24.}\label{fig:EM}
 \end{figure*}


 \begin{figure*}
\centering
 \includegraphics[width=.45\textwidth]{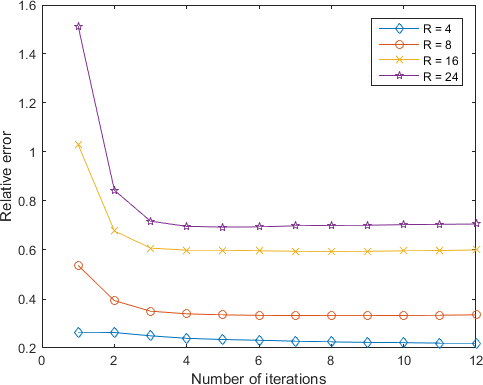}\qquad
 \includegraphics[width=.45\textwidth]{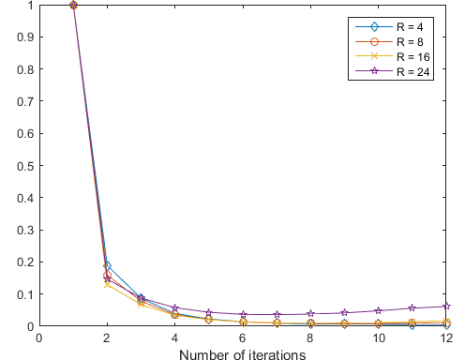}
\caption{\textbf{Left:} Relative error per iteration of the proposed 3DShearCS method for different undersampling factors. \textbf{Right:} Relative changes of the weights at each iteration level. Already after three iterations the relative change of the weights is less than 10\%.}\label{fig:Iter3DAndWeights}
 \end{figure*}
 
 The quantitative image quality parameters vessel sharpness, HaarPSI, and the relative error are shown in Figure \ref{fig:EM} and confirm the higher image quality of 3D compared to 2D shearlet-based CS reconstruction.
%

Figure \ref{fig:Iter3DAndWeights} shows the relative error of the reconstruction at each iteration. It can be observed that for all undersampling factors R = 4, 8, 16 and 24 the algorithm converges after as few as four iterations. This fast convergence is also reflected in the relative change of the weights (right picture in Figure \ref{fig:Iter3DAndWeights}). During the first few iterations the weights do change significantly but already after three or four iterations the relative change of the weights is less than 10\%. The weights only change significantly if the image update in the reconstruction change significantly. Hence, it is not surprising that for higher undersampling rates, such as R = 24, the weights change more than, for example for R = 4. It is also important to mention that we have not changed the parameters $\beta$ and $\mu$ for the different undersampling rates which is usually needed for other methods that do 
not involve any adaptivity. 
 
Based on these results we keep the weights fixed after three iterations and limit the total number of iterations to 12 to minimize reconstruction times.


\subsection{In Vivo Evaluation}
\begin{figure*}[h!]
\centering
\qquad  \quad
\begin{overpic}[width=0.9\textwidth]{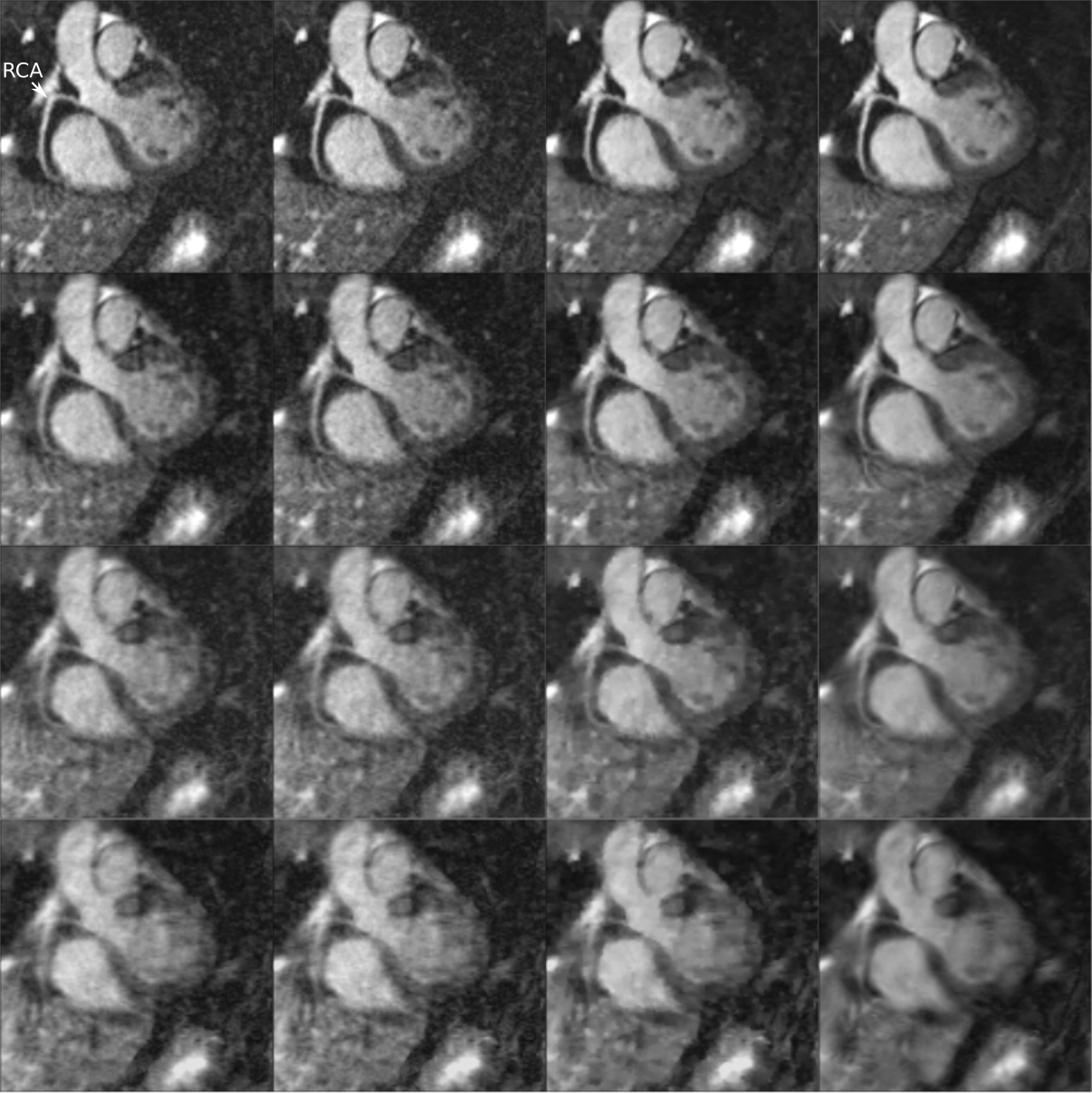} 
 \put(10,100.5){itSense}
 \put(35,100.5){WaveCS}
 \put(62,100.5){TV}
 \put(83,100.5){3DShearCS}
\put(-8,86.5){R = 4}
 \put(-8,61.5){R = 8}
 \put(-8,36.5){R = 16}
 \put(-8,11.5){R = 24}
\end{overpic}
\caption{Reformatted images showing the right coronary artery (RCA) in a healthy volunteer for different undersampling factors and different reconstruction technqiues. The proposed 3DShearCS approach ensures high image quality and good anatomical depiction even for undersampling factors as high as 24, corresponding to a scan time of 1.6 min for a 3D whole-heart scan with isotropic resolution of 1.5 $mm^3$.}\label{fig:InVivo}
\end{figure*}

Fig. \ref{fig:InVivo}  depicts reformatted images of one volunteer showing the right and left coronary artery. Higher undersampling factors lead to a higher degree of incoherent undersampling artefacts. For moderate degrees of undersampling WaveCS yields high image quality but for R $>$ 4 it does not perform significantly better than standard itSENSE. TV is more robust towards undersampling artefacts but for R = 16 regularisation artefacts become visible which make the anatomy appear to have jagged rather than smooth edges. 3DShearCS on the other hand yields anatomically more accurate depiction of the cardiac anatomy and coronary arteries even for scan times as short as 1.6 min (R = 24). 

\begin{figure*}[h!]
\centering
\includegraphics[width = \textwidth]{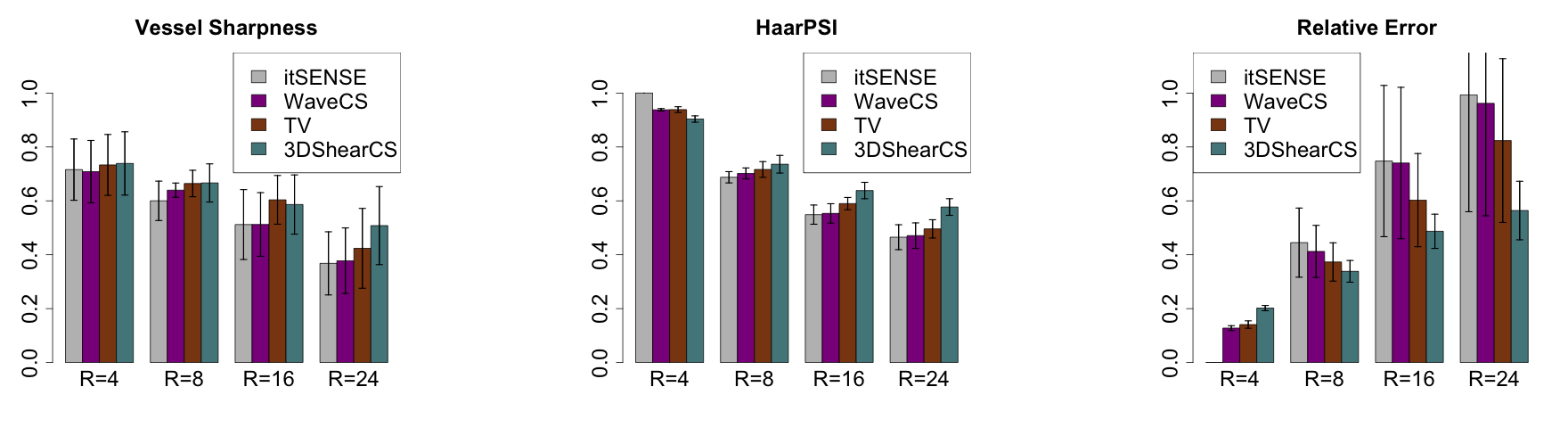}
 \caption{Evaluation of in-vivo image quality using vessel sharpness, HaarPSI and relative error. All three metrics show that the proposed 3DShearCS approach is more robust towards undersampling. It is important to note that for vessel sharpness and HaarPSI high values indicate good image quality whereas for the relative error low values are more favorable. Values are depicted as mean and standard deviation over five volunteers. Note that the HaarPsi is 1 and the relative error is 0 for itSENSE and R = 4 as this is the reference image.
 }\label{fig:Barplots}
\end{figure*}

The results of the quantitative assessment of the image quality are summarized in Fig. \ref{fig:Barplots} and are in agreement with the above assessment. For R = 4 and R = 8 all reconstruction methods perform comparably well, for R = 16 WaveCS performs worse than 3DShearCS and TV but yields similar results as itSENSE. For R = 24 3DShearCS leads to superior image quality than the other approaches with improvements in vessel sharpness of 42$\pm$28\% (p = 0.014) compared to itSENSE,  39$\pm$34\% (p = 0.022) compared to WaveCS and 23$\pm$24\% (p = 0.033) compared to TV. HaarPsi and relative error show similar behavior with a reduction in RE of 37$\pm$19\% compared to itSENSE (p = 0.004), 35$\pm$18\% (p = 0.005) compared to WaveCS, 27$\pm$15\% (p = 0.002) compared to TV and an improvement of 25$\pm$11\% (p = 0.042) in HaarPsi compared to itSENSE,23$\pm$11\% (p = 0.045) compared to WaveCS, 17$\pm$6\% (p = 0.041) compared to TV. 

The clinical experts assessed 20 different reconstructions (five volunteers x four different undersampling factors) and compared the four different reconstruction methods to each other and to a reference image which was chosen as itSENSE with R = 4 (Table \ref{table:Rev1} and \ref{table:Rev2}). Reviewer 1 and Reviewer 2 scored a similar number of reconstructions to be of diagnostic quality, with non-diagnostic images occurring only for R = 24 and for R = 16. The proposed 3DShearCS approach was scored with the highest image score by both reviewers, and images reconstructed with 3DShearCS agreed best with the reference image in the majority of cases. 

\begin{table}[H]
\centering
\begin{tabularx}{.5\textwidth}{|X|X|X|X|X|}
\hline
 & itSENSE  &  WaveCS  &  TV  &  3DShearCS \\
\hline
Diagnostic & 16/20 & 17/20 & 16/20 & 16/20 \\
\hline
Image score & 1.15 & 1.15 & 1.2 & 1.2 \\
\hline
Agreement & 5/20 & 7/20 & 3/20 & 15/20 \\
\hline
\end{tabularx}
\caption{Image evaluation from clinical reviewer 1.}\label{table:Rev1}
\end{table}

\begin{table}[H]
\centering
\begin{tabularx}{.5\textwidth}{|X|X|X|X|X|}
\hline
 & itSENSE  &  WaveCS  &  TV  &  3DShearCS \\
\hline
Diagnostic & 15/20 & 16/20 & 17/20 & 18/20 \\
\hline
Image score & 1.0 & 1.05 & 1.05 & 1.4 \\
\hline
Agreement & 0/20 & 1/20 & 5/20 & 14/20 \\
\hline
\end{tabularx}
\caption{Image evaluation from clinical reviewer 2.}\label{table:Rev2}
\end{table}

\section{Discussion}
The proposed 3DShearCS approach was demonstrated to yield a better image quality than the reference methods. We have shown that extending the shearlet system from 2D to 3D leads to a better suppression of undersampling artefacts. The proposed iterative reweighting scheme ensures fast convergence and high image quality. 

Relative error and HaarPSI are commonly used image metrics but they only describe the difference to a reference image, without providing any information if the measured difference improves or decreases the image quality. This limitation can be seen in Fig. \ref{fig:Barplots} for R = 4. The reference image is itSENSE with R = 4 and HaarPSI and the relative error suggest that TV, WaveCS and 3DShearCS perform worse than itSENSE for R = 4. Nevertheless, the images in Fig. \ref{fig:InVivo} and the vessel sharpness in Fig. \ref{fig:Barplots} indicate that especially TV and 3DShearCS lead to a better image quality than itSENSE. 

The performance of 3DShearCS depends on the resolution of the images. Optimal results for a 3D shearlet system can only be achieved with high isotropic resolution and isotropic number of pixels. Here we used a 3D non-Cartesian trajectory which is specifically designed to fulfill this requirement. For other MR sampling schemes which provide anisotropic image data, for instance if the dimension in one direction is significantly smaller than the other two and those are very large, a 2D shearlet approach should be considered. 

Another optimization option for the reconstruction is the shearlet filter. Depending on the data and the experimental setup the filter can be adapted to the image content. For instance, in vessel wall imaging more focus could be put on the edges, hence, smaller filters with more spatial localization may perform better than filters with large support. 

The proposed reweighting approach could also be applied to 3D wavelet transforms, but only a 3D redundant wavelet transform would significantly benefit from the reweighting scheme which, to the best of our knowledge, is not available for Matlab. In addition, previously proposed WaveCS approaches do not use 3D redundant wavelet transform and therefore we compared 3DShearCS to a wavelet-based CS without iterative reweighting. 

One of the main challenges of advanced MR image reconstruction approaches are reconstruction times. For this study the reconstruction times were 4 hours on a 2 x Intel Xeon X2630v2  Hexa-Core machine with 64 GB memory. However, the code is also not optimized for speed but will be as a part of future work.

All the image reconstruction techniques used in this work required data dependent reconstruction parameters which had to be manually set. For 3DShearCS the same set of reconstruction parameters was used for all in-vivo experiments. The TV based-reconstruction required a change of parameters for different undersampling factors.

\section{Conclusion}
We have presented a novel CS reconstruction approach using shearlet-based sparsity transforms which can be used for arbitrary MR sampling patterns and exploits the information from multiple receiver coils to improve the conditioning of the reconstruction problem. An iterative reweighting approach is used to ensure fast convergence of the algorithm and high image quality even for a low number of iterations, improving both the accuracy of the approach while reducing reconstruction times. The proposed 3DShearCS method was evaluated in 3D in-vivo cardiac MR scans of healthy volunteers. Compared to existing reconstruction methods, 3DShearCS lead to superior image quality which was determined using standard image quality metrics and assessments of clinical experts. 3DShearCS ensures high quality 3D MR images with high isotropic resolution in short scan times and could help to promote 3D high-resolution CMR in clinical practice.  

\section*{Acknowledgements}
This work was done when J. Ma was affiliated with the Technische Universit\"{a}t Berlin, he acknowledges support from the DFG Collaborative Research Center TRR 109 ``Discretization in Geometry and Dynamics'' and the Berlin Mathematical School.
G. Kutyniok acknowledges partial support by the Einstein Foundation Berlin, the Einstein Center for Mathematics Berlin (ECMath), the European Commission-Project DEDALE (contract no. 665044) within the H2020 Framework Program, DFG Grant KU 1446/18, DFG-SPP 1798 Grants KU 1446/21 and KU 1446/23, the DFG
Collaborative Research Center TRR 109 Discretization in Geometry and Dynamics, and by the DFG Research Center {\sc Matheon} ÔMathematics for Key TechnologiesÕ in Berlin.

\ifCLASSOPTIONcaptionsoff
  \newpage
\fi

\bibliographystyle{IEEEtran}

\bibliography{IEEEabrv,./bib}

\end{document}